\documentclass[a4paper,twoside]{article}

\usepackage{epsfig}
\usepackage{subcaption}
\usepackage{calc}
\usepackage{amssymb}
\usepackage{amstext}
\usepackage{amsmath}
\usepackage{amsthm}
\usepackage{multicol}
\usepackage{pslatex}
\usepackage{apalike}
\usepackage{algorithm2e}
\usepackage[bottom]{footmisc}

\usepackage[table]{xcolor}
\usepackage{euscript}
\usepackage{amssymb}
\usepackage{multirow}
\usepackage[misc]{ifsym}
\usepackage[super]{nth}
\usepackage[hidelinks]{hyperref}

\definecolor{tabseccolor}{rgb}{0.0, 0.0, 1.0}
\newcommand{\tabseccol}[1]{\textcolor{tabseccolor}{#1}}
\definecolor{tabthicolor}{rgb}{0.9, 0.5, 0.0}
\newcommand{\tabthicol}[1]{\textcolor{tabthicolor}{#1}}

\definecolor{tabbettercolor}{rgb}{0.0, 0.9, 0.0}
\newcommand{\tabbetter}[1]{\textcolor{tabbettercolor}{\textbf{\underline{#1}}}}
\definecolor{tabworsecolor}{rgb}{0.9, 0.0, 0.0}
\newcommand{\tabworse}[1]{\textcolor{tabworsecolor}{\textbf{#1}}}

\newcommand{\tabbest}[1]{\textbf{#1}}
\newcommand{\tabsec}[1]{\tabseccol{\underline{#1}}}
\newcommand{\tabthi}[1]{\tabthicol{\textit{#1}}}

\usepackage{xcolor,colortbl}

\newcommand{\lightvalue}{0.92}
\definecolor{lightgray}{rgb}{\lightvalue,\lightvalue,\lightvalue}
\definecolor{lightgreen}{rgb}{\lightvalue,1,\lightvalue}
\definecolor{lightred}{rgb}{1,\lightvalue,\lightvalue}
\definecolor{lightblue}{rgb}{\lightvalue,\lightvalue,1}
\definecolor{lightyellow}{rgb}{1,1,\lightvalue}

\newcommand{\midvalue}{0.84}
\definecolor{midgreen}{rgb}{\midvalue,1,\midvalue}

\usepackage{arydshln}

\usepackage{dcolumn}        
\newcolumntype{C}{D{.}{.}{2.2}} 

\usepackage{glossaries}

\newacronym{scn}{SCN}{Spatial-ConfigurationNet}
\newacronym{nnunet}{nnU-Net}{no-new-Net}
 
\newacronym{gdloss}{GD loss}{Generalized Dice Loss}
\newacronym{celoss}{CE loss}{Cross-Entropy Loss}

\newacronym[plural=CNNs,firstplural=Convolutional Neural Networks (CNN)]{cnn}{CNN}{Convolutional Neural Network}
\newacronym[plural=ASMs,firstplural=Active Shape Models (ASM)]{asm}{ASM}{Active Shape Model}
\newacronym[plural=AAMs,firstplural=Active Appearance Models (AAM)]{aam}{AAM}{Active Appearance Model}
\newacronym{cca}{CCA}{connected component analysis}
\newacronym{mas}{MAS}{Multi-Atlas Segmentation}
\newacronym{mcd}{MCD}{Monte-Carlo Dropout}
\newacronym[plural=CaRe-CNNs,firstplural=Cascading Refinement CNNs (CaRe-CNN)]{carecnn}{CaRe-CNN}{Cascading Refinement CNN}

\newacronym{iid}{i.i.d.}{independent and identically distributed}
\newacronym{tl}{TL}{Transfer Learning}
\newacronym{da}{DA}{Domain Adaptation}
\newacronym{dg}{DG}{Domain Generalization}
\newacronym{sfda}{SFDA}{Source-Free Domain Adaptation}
\newacronym{tta}{TTA}{Test-time Adaptation}

\newacronym{ct}{CT}{computed tomography}
\newacronym{mr}{MR}{magnetic resonance}
\newacronym{lge}{LGE}{late gadolinium enhanced}
\newacronym{hu}{HU}{Hounsfield Units}

\newacronym{miccai}{MICCAI}{International Conference on Medical Image Computing and Computer-Assisted Intervention}
\newacronym{mmwhs}{MMWHS}{Multi-Modality Whole Heart Segmentation}
\newacronym{flare}{FLARE}{Fast and Low GPU Memory Abdominal Organ Segmentation}
\newacronym{amos}{AMOS}{Multi-Modality Abdominal Multi-Organ Segmentation}
\newacronym{bcv}{BCV}{Multi-Atlas Labeling Beyond the Cranial Vault}
\newacronym{bcvc}{BCV-C}{BCV-Cervix}
\newacronym{bcva}{BCV-A}{BCV-Abdomen}
\newacronym{myosaiq}{MYOSAIQ}{Myocardial Segmentation with Automated Infarct Quantification}
\newacronym{fimh}{FIMH}{International Conference on Functional Imaging and Modeling of the Heart}
\newacronym{emidec}{EMIDEC}{\textit{Automatic Evaluation of Myocardial Infarction from Delayed-Enhancement Cardiac MRI}}

\newacronym[plural=ROIs,firstplural=regions of interest (ROI)]{roi}{ROI}{region of interest}
\newacronym[plural=GPUs,firstplural=graphics processing units (GPU)]{gpu}{GPU}{graphics processing unit}

\newacronym{mi}{MI}{myocardial infarction}

\newacronym{lv}{LV}{left ventricle cavity}
\newacronym{myo}{MYO}{healthy myocardium}
\newacronym{mit}{MIT}{myocardial infarct tissue}
\newacronym{mvo}{MVO}{microvascular obstruction}
\newacronym{fmyo}{f-MYO}{full myocardium}
\newacronym{fmit}{f-MIT}{full myocardial infarct tissue}

\newacronym{cg}{CG}{central gland}
\newacronym{pz}{PZ}{peripheral zone}

\newacronym{li}{LI}{liver}
\newacronym{lk}{LK}{left kidney}
\newacronym{rk}{RK}{right kidney}
\newacronym{sp}{SP}{spleen}
\newacronym{st}{ST}{stomach}

\newacronym{bmc}{BMC}{Boston Medical Center}
\newacronym{runmc}{RUNMC}{Radboud University Nijmegen Medical Center}

\newacronym{dsc}{DSC}{Dice score}
\newacronym{hd}{HD}{Hausdorff distance}
\newacronym{assd}{ASSD}{average symmetric surface distance}
\newacronym{cc}{CC}{correlation coefficient score}
\newacronym{mae}{MAE}{mean absolute error}
\newacronym{loa}{LOA}{limits of agreement}
\newacronym{crps}{CRPS}{continuous ranked probability score}

\newacronym{psir}{PSIR}{phase sensitive inversion recovery}
\newacronym{mag}{MAG}{magnitude reconstruction}
\newacronym{bmm}{BMM}{boundary mining model}
\newacronym{alm}{ALM}{adversarial learning model}

\newacronym{cdcblock}{CDC block}{convolution-dropout-convolution block}

\newcommand{\lossbase}{L}
\newcommand{\lossdice}{\lossbase_{\text{GD}}}

\newcommand{\imagelettersmall}{x}
\newcommand{\groundtruthlettersmall}{y}

\newcommand{\predictionlettersmall}{p}
\newcommand{\finalpredlettersmall}{f}

\newcommand{\image}{\mathbf{\imagelettersmall}}
\newcommand{\groundtruth}{\mathbf{\groundtruthlettersmall}}

\newcommand{\prediction}{\mathbf{\hat{\groundtruthlettersmall}}}

\newcommand{\predlettersmall}{\mathbf{\hat{\predictionlettersmall}}}
\newcommand{\predstageone}{\predlettersmall_1}
\newcommand{\predstagetwo}{\predlettersmall_2}
\newcommand{\predstagethree}{\predlettersmall_3}

\newcommand{\labelpredstageone}{\mathbf{\hat{\groundtruthlettersmall}}_1}
\newcommand{\labelpredstagetwo}{\mathbf{\hat{\groundtruthlettersmall}}_2}
\newcommand{\labelpredstagethree}{\mathbf{\hat{\groundtruthlettersmall}}_3}
\newcommand{\labelpredfinal}{\mathbf{\hat{\groundtruthlettersmall}}_\finalpredlettersmall}

\newcommand{\gtstageone}{\mathbf{\groundtruthlettersmall}_1}
\newcommand{\gtstagetwo}{\mathbf{\groundtruthlettersmall}_2}
\newcommand{\gtstagethree}{\mathbf{\groundtruthlettersmall}_3}

\newcommand{\modelweightsbase}{\theta}
\newcommand{\modelweightsstageone}{{\modelweightsbase}_1}
\newcommand{\modelweightsstagetwo}{{\modelweightsbase}_2}
\newcommand{\modelweightsstagethree}{{\modelweightsbase}_3}

\newcommand{\modelbase}{\mathcal{M}}
\newcommand{\modelstageone}{{\modelbase}_1}
\newcommand{\modelstagetwo}{{\modelbase}_2}
\newcommand{\modelstagethree}{{\modelbase}_3}

\newcommand{\lossfactorbase}{\lambda}
\newcommand{\lossfactorone}{{\lossfactorbase}_1}
\newcommand{\lossfactortwo}{{\lossfactorbase}_2}
\newcommand{\lossfactorthree}{{\lossfactorbase}_3}

\newcommand{\checkx}{×}

\usepackage{SCITEPRESS}     

\begin{document}

\title{CaRe-CNN: Cascading Refinement CNN for Myocardial Infarct Segmentation with Microvascular Obstructions}

\author{\authorname{Franz Thaler\sup{1,2}\orcidAuthor{0000-0002-6589-6560}, Matthias A.F. Gsell\sup{1}\orcidAuthor{0000-0001-7742-8193}, Gernot Plank\sup{1}\orcidAuthor{0000-0002-7380-6908} and Martin Urschler\sup{3}\orcidAuthor{0000-0001-5792-3971}}
\affiliation{\sup{1}Gottfried Schatz Research Center: Medical Physics and Biophysics, Medical University of Graz, Graz, Austria}
\affiliation{\sup{2}Institute of Computer Graphics and Vision, Graz University of Technology, Graz, Austria}
\affiliation{\sup{3}Institute for Medical Informatics, Statistics and Documentation, Medical University of Graz, Graz, Austria}
}


\abstract{
Late gadolinium enhanced (LGE) magnetic resonance (MR) imaging is widely established to assess the viability of myocardial tissue of patients after acute myocardial infarction (MI).
We propose the Cascading Refinement CNN (CaRe-CNN), which is a fully 3D, end-to-end trained, 3-stage CNN cascade that exploits the hierarchical structure of such labeled cardiac data.
Throughout the three stages of the cascade, the label definition changes and CaRe-CNN learns to gradually refine its intermediate predictions accordingly.
Furthermore, to obtain more consistent qualitative predictions, we propose a series of post-processing steps that take anatomical constraints into account.
Our CaRe-CNN was submitted to the FIMH 2023 MYOSAIQ challenge, where it ranked second out of 18 participating teams.
CaRe-CNN showed great improvements most notably when segmenting the difficult but clinically most relevant myocardial infarct tissue (MIT) as well as microvascular obstructions (MVO).
When computing the average scores over all labels, our method obtained the best score in eight out of ten metrics.
Thus, accurate cardiac segmentation after acute MI via our CaRe-CNN allows generating patient-specific models of the heart serving as an important step towards personalized medicine.
}

\onecolumn \maketitle \normalsize \setcounter{footnote}{0} \vfill

\section{\uppercase{Introduction}}
\label{sec:introduction}

\begin{figure*}[t] 
\includegraphics[width=\textwidth]{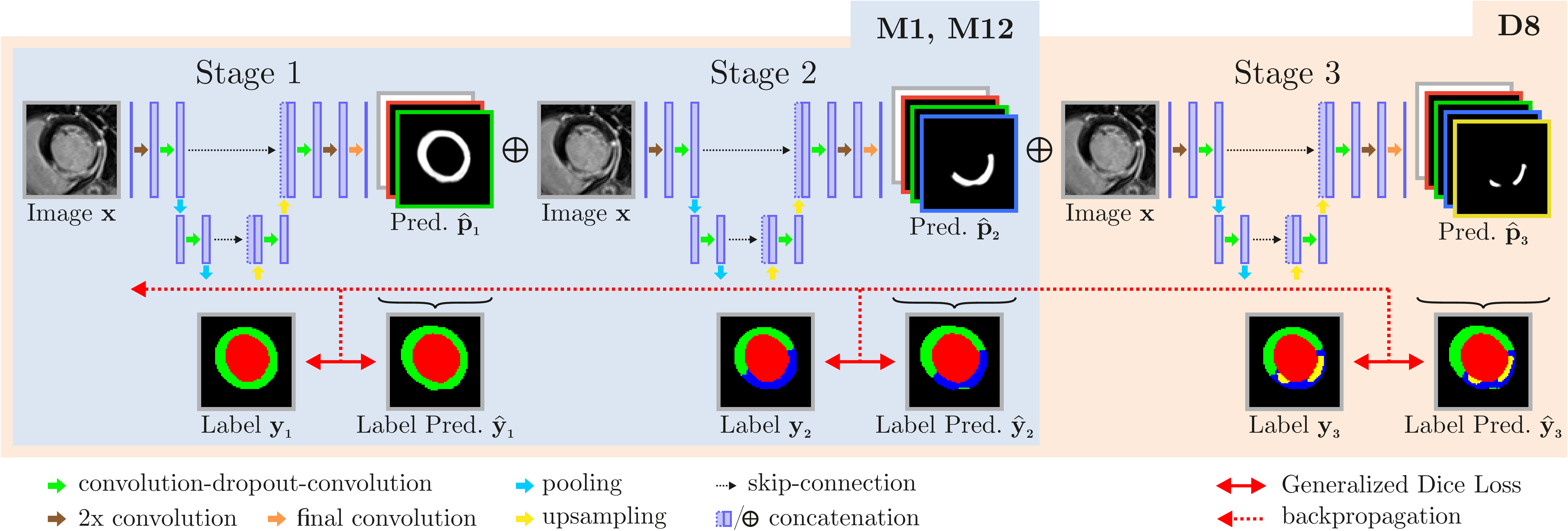}
\caption{
Overview of the proposed CaRe-CNN architecture for segmenting cardiac LGE MR images after MI.
CaRe-CNN is a 3-stage CNN cascade that exploits the hierarchical label definition of the data and refines intermediate predictions in consecutive stages.
The whole architecture is trained end-to-end and all data is processed in 3D.
As MVO can only be present for data of the D8 subgroup, we consider Stage~2 predictions as final predictions for data of the M1 and M12 subgroups.
}
\label{fig:overview}
\end{figure*}

Cardiovascular diseases are the leading cause of death worldwide among which \gls{mi} is one of the most prevalent diseases\footnote{\href{https://www.who.int/health-topics/cardiovascular-diseases}{https://www.who.int/health-topics/cardiovascular-diseases}, last accessed on October 8, 2023}.
\gls{mi} is caused by a decrease or complete cessation of blood flow in the coronary arteries 
which reduces perfusion in the supplied myocardial tissue, 
leading to a metabolic undersupply that impairs cardiac function and, ultimately, 
may result in myocardial necrosis.
The accurate assessment of tissue damage after acute \gls{mi} is highly relevant 
as the extension of myocardial necrosis is an important risk factor for developing heart failure.
On one hand, viable myocardial tissue  
with a potential for functional recovery on restoration of normal blood supply by revascularization might recover~\cite{wroblewski1990evaluation,perin2002assessing}, 
which may improve the functional capacity and survival~\cite{van1996magnetic,kim2004viability}.
On the other hand, precise delineation of infarcted myocardial tissue is crucial 
to determine the risk of further adverse cardiovascular events like ventricular tachycardia 
which may lead to sudden death~\cite{rosenthal1985sudden,hellermann2002heart}.
For example, the presence of microvascular obstructions, characterized by a damaged microvasculature resulting in a 'no-reflow' phenomenon preventing blood flow from penetrating beyond the myocardial capillary bed, is linked to adverse ventricular remodeling and an increased risk of future cardiovascular events~\cite{hamirani2014effect,rios2019microvascular}.
Thus, the accurate assessment of post-\gls{mi} tissue damage is of pivotal importance.
In clinical practice \gls{mr} imaging is used to quantify areas of impaired myocardial function e.g.\ by estimating the end-diastolic wall thickness of the left ventricle, or by evaluating the contractile reserve, i.e. the myocardial stress-to-rest ratio~\cite{kim1999relationship,schinkel2007assessment}.
One of the most accurate methods is \gls{lge} \gls{mr} imaging, where the contrast agent accumulates in impaired tissue areas, thus allowing to visualize the transmural extent of tissues affected by \gls{mi}~\cite{selvanayagam2004value}.

However, analyzing \gls{lge} \gls{mr} images to characterize tissue viability in an accurate and efficient manner 
remains a significant challenge.  
Nowadays, deep learning-based \glspl{cnn} are widely adopted to medical image analysis tasks like the detection of diseases in medical images~\cite{Esteva2017,Feng2022-md}, or image segmentation of the brain~\cite{akkus2017deep}, the vertebrae~\cite{Payer2020-yv}, or the heart~\cite{chen2020deep}.
From cardiac \gls{lge} \gls{mr} data, healthy and necrotic myocardial tissue can be assessed by \gls{cnn}-based medical image segmentation, where each voxel of an \gls{lge} \gls{mr} image is assigned the respective label.
Accurate cardiac segmentation of patients after \gls{mi} can provide a foundation for generating anatomically accurate
patient-specific models of the heart, which, in turn, can be used e.g., to create cardiac digital twin models of
human electrophysiology \cite{gillette2021framework} to identify potential patient-specific causes for arrhythmia 
improving personalized therapy planing \cite{Campos2022}.

Due to the challenging nature of fully automated infarct segmentation, some approaches in the literature rely on manual segmentations of the full myocardium such that a distinction between healthy and infarcted tissue only needs to be learned within that region~\cite{zabihollahy2018myocardial,moccia2019development}.
Instead of using \gls{lge} \gls{mr} data, \cite{xu2018direct} uses cine \gls{mr} data without contrast agents and a Long Short-Term Memory-based Recurrent Neural Network~\cite{graves2013speech} to predict myocardial infarct tissue from motion.
In contrast to that, \cite{fahmy2018automated} automatically segment both, healthy and infarcted tissue from \gls{lge} \gls{mr} images by employing a 2D \gls{cnn} based on the U-Net~\cite{ronneberger2015u} architecture.
In another fully-automated segmentation approach, \cite{chen2022automatic} employed two consecutive 2D U-Net-like \glspl{cnn} as a cascade, where the first network learns to segment the full myocardium, while the second is trained to refine the prediction to obtain the infarct region.
The authors show that the consecutive setup achieves better Dice and Jaccard scores, but worse volume estimation compared to a parallel setup of two \glspl{cnn}.
The semi-supervised myocardial infarction segmentation approach in~\cite{xu2022bmanet} proposes to use attention mechanisms to obtain the coarse location of the myocardial infarction before refining the prediction step-by-step.
In order to allow training from unlabeled data, they use an adversarial learning model that provides a training objective even when ground truth labels are not available.
The EMIDEC challenge held in conjunction with the \gls{miccai} in 2020 aimed to automatically segment myocardial infarct regions from \gls{lge} \gls{mr} images in their segmentation track~\cite{lalande2022deep}.
Different one- and two-stage approaches mostly based on U-Net-like architectures were submitted by the challenge participants.
The highest scores in the segmentation track were achieved by \cite{zhang2021cascaded} who employed a coarse to fine two-stage approach, where initial predictions are obtained from a 2D U-Net variant before all 2D predictions are stacked to a 3D volume.
The stacked prediction in combination with the \gls{lge} \gls{mr} image is then refined by a 3D U-Net variant to obtain the final prediction.

In this work, we propose the \gls{carecnn}, which -- differently to related work -- is a fully 3D, end-to-end trained 3-stage \gls{cnn} cascade that exploits the hierarchical structure of cardiac \gls{lge} \gls{mr} images after \gls{mi} and sequentially refines the predicted segmentations.
Further, we propose a series of post-processing steps that take anatomical constraints into account to obtain more consistent qualitative predictions.
Our \gls{carecnn} was submitted to the \gls{myosaiq} challenge which was held in conjunction with the \gls{fimh} 2023.
We evaluate our method by comparing to state-of-the-art methods submitted to the \gls{myosaiq} challenge where our \gls{carecnn} ranked second out of 18 participating teams.

\section{\uppercase{Method}}

In this work we propose \gls{carecnn}, a cascading refinement \gls{cnn} to semantically segment different cardiac structures after \gls{mi} from \gls{lge} \gls{mr} images in 3D.
An overview of \gls{carecnn} is provided in Fig.~\ref{fig:overview}.

\subsection{Notation and Definitions}

\begin{figure}
\includegraphics[width=0.475\textwidth]{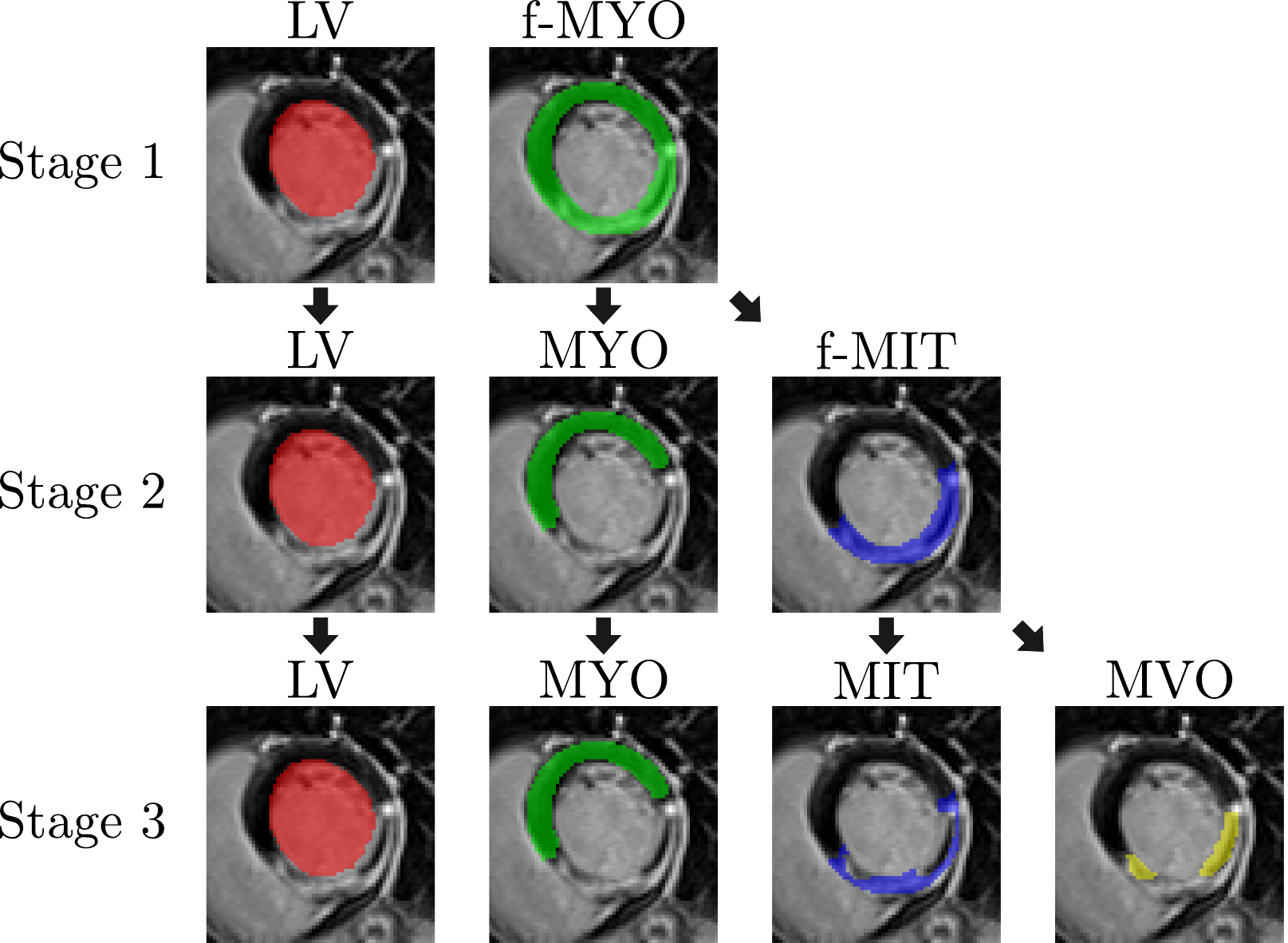}
\caption{
Visualization of the hierarchical label definitions per stage as used by \gls{carecnn}.
While \gls{lv} remains unchanged, \gls{fmyo} can be separated into \gls{myo} and \gls{fmit} of which the latter can be separated into \gls{mit} and \gls{mvo}.
}
\label{fig:labels_per_stage}
\end{figure}

Throughout this work, we will refer to the labels as \gls{lv}, \gls{myo}, \gls{mit} and \gls{mvo}.
For further disambiguation of intermediate results at the different stages of our method, we additionally define the \gls{fmyo} as \gls{myo} $\cup$ \gls{mit} $\cup$ \gls{mvo} and the \gls{fmit} as \gls{mit} $\cup$ \gls{mvo}.
A visualization of the label definitions at different stages is provided in Fig.~\ref{fig:labels_per_stage}.
While all scans in the dataset are \gls{lge} \gls{mr} images after \gls{mi}, the dataset can be split into three subgroups (D8, M1, M12) depending on how much time has passed since the \gls{mi}, see Section~\ref{sec:dataset}.
Importantly, \gls{mvo} is exclusive to the D8 subgroup and the subgroup information is well-known for every image in the training and test set.

\subsection{Cascading Refinement CNN}

Our \gls{carecnn} architecture exploits the hierarchical structure of the semantic labels and is set up as a cascade of three consecutive 3D U-Net-like architectures~\cite{ronneberger2015u} which are trained end-to-end.
Throughout this work, we will refer to each of these consecutive parts of the processing pipeline as stages numbered from 1 to 3.
By design, any subsequent stage of \gls{carecnn} receives the prediction of the preceding stage as additional input, such that the prediction is gradually refined, see Fig.~\ref{fig:overview}.

After randomly choosing and preprocessing a 3D image $\image$ with ground truth $\groundtruth$ from the training set, the image $\image$ is provided as input to \gls{carecnn}.
Stage~1 of \gls{carecnn} aims to distinguish between the \gls{lv}, the \gls{fmyo} and the background based on the image information.
By denoting the Stage~1 model as $\modelstageone(\cdot)$ with trainable parameters $\modelweightsstageone$, the output prediction $\predstageone$ of Stage~1 for image $\image$ can be expressed as:
\begin{equation}
    \predstageone = \modelstageone (\image ; \modelweightsstageone).
\end{equation}
Please note that the output prediction $\predlettersmall$ refers to the model output without activation function.
In Stage~2, \gls{carecnn} learns to predict the \gls{lv}, the healthy \gls{myo}, the \gls{fmit} and the background by refining the Stage~1 prediction.
To allow consecutive refinement of $\predstageone$ in Stage~2, we provide $\predstageone$ concatenated with the original image $\image$ in the channel dimension as input to the Stage~2 model.
This way, $\predstageone$ can be refined based on the original image information which is crucial for our cascading \gls{cnn} as the label definition of the individual stages is not the same.
The Stage~2 model $\modelstagetwo(\cdot)$ with trainable parameters $\modelweightsstagetwo$ is defined as: 
\begin{equation}
    \predstagetwo = \modelstagetwo (\predstageone \oplus \image ; \modelweightsstagetwo),
\end{equation}
where $\oplus$ refers to a concatenation in the channel dimension and $\predstagetwo$ refers to the output prediction of Stage~2, again without any activation function.
Lastly, Stage~3 aims to distinguish all labels, i.e., the \gls{lv}, \gls{myo}, \gls{mit}, \gls{mvo} as well as the background.
To continue our \gls{cnn} cascade, we concatenate the prediction $\predstagetwo$ and the image $\image$ in the channel dimension to provide both as input to the Stage~3 model $\modelstagethree(\cdot)$ of our cascading \gls{cnn}.
Formally, the output prediction $\predstagethree$ of Stage~3 can be expressed as:
\begin{equation}
    \predstagethree = \modelstagethree (\predstagetwo \oplus \image ; \modelweightsstagethree),
\end{equation}
where $\modelweightsstagethree$ refers to the trainable parameters of Stage~3 and $\oplus$ defines the concatenation operator.

\subsection{Training Objective}

In our training pipeline the segmentation loss is computed for each stage individually and backpropagation through all stages is allowed to update model weights in an end-to-end manner for the whole cascade.
As the label definition varies from stage to stage, we adapt the ground truth labels such that they follow the label definition of the respective stage as defined in Fig.~\ref{fig:labels_per_stage}. 
For every stage, we compute the generalized Dice loss between the ground truth $\groundtruth$ and the label prediction $\prediction = \text{softmax}  ( \predlettersmall )$ of that stage.
Formally, the generalized Dice loss $\lossdice(\cdot)$ is expressed as:
\begin{equation}
    \lossdice(\groundtruth, \prediction) = 1 - 2 \frac{\sum_{k=1}^{K}  w_k \cdot \sum_{m=1}^M \prediction_{m} \cdot \groundtruth_{m}}{\sum_{k=1}^{K} w_k \cdot \sum_{m=1}^M \prediction_{m}^2 + \groundtruth_{m}},
\label{eq:loss_function_general}
\end{equation}
where $K$ represents the number of all labels and $M$ is the number of voxels.
The label weight $w_k$ for label $k$ is computed as the ratio of voxels $M_k$ with label $k$ in the ground truth compared to the number of all voxels, i.e. $w_k = \frac{M_k}{M}$.
The square term $\prediction_{m}^2$ is used to account for class imbalance.

During training only images that actually contain the \gls{mvo} label are forwarded through Stage~3 as images with missing labels might lead to unstable training which can greatly impact the performance at that stage.
In order to provide a loss at every stage for every iteration while also allowing all training images to be selected at some point, we always randomly pick two training images per iteration: One image with and one without the \gls{mvo} label.
The overall training objective of \gls{carecnn} for all stages and a single image can then be expressed as:
\begin{equation}
\begin{split}
\lossbase(\groundtruth, \prediction) &=
\lossfactorone
\underbrace{\lossdice(\gtstageone, \labelpredstageone; \modelweightsstageone)}_{\text{update $\modelstageone$}}
+ \lossfactortwo
\underbrace{\lossdice(\gtstagetwo, \labelpredstagetwo; \modelweightsstageone, \modelweightsstagetwo)}_{\text{update $\modelstageone$ and $\modelstagetwo$}}
\\ &+ \delta_{\text{MVO}} \cdot \lossfactorthree
\underbrace{\lossdice(\gtstagethree, \labelpredstagethree; \modelweightsstageone, \modelweightsstagetwo, \modelweightsstagethree)}_{\text{update $\modelstageone$, $\modelstagetwo$ and $\modelstagethree$}},
\end{split}
\label{eq:mlc_loss}
\end{equation}
where the stage weights $\lossfactorone$, $\lossfactortwo$ and $\lossfactorthree$ serve as weights between the individual loss terms and are set to 1.
The term $\delta_{\text{MVO}}$ is set to $1$ if ground truth $\groundtruth$ contains \gls{mvo} anywhere and is $0$ otherwise.
Finally, we provide the mean loss over the batch to the optimizer.

\begin{figure}[t] 
\includegraphics[width=0.475\textwidth]{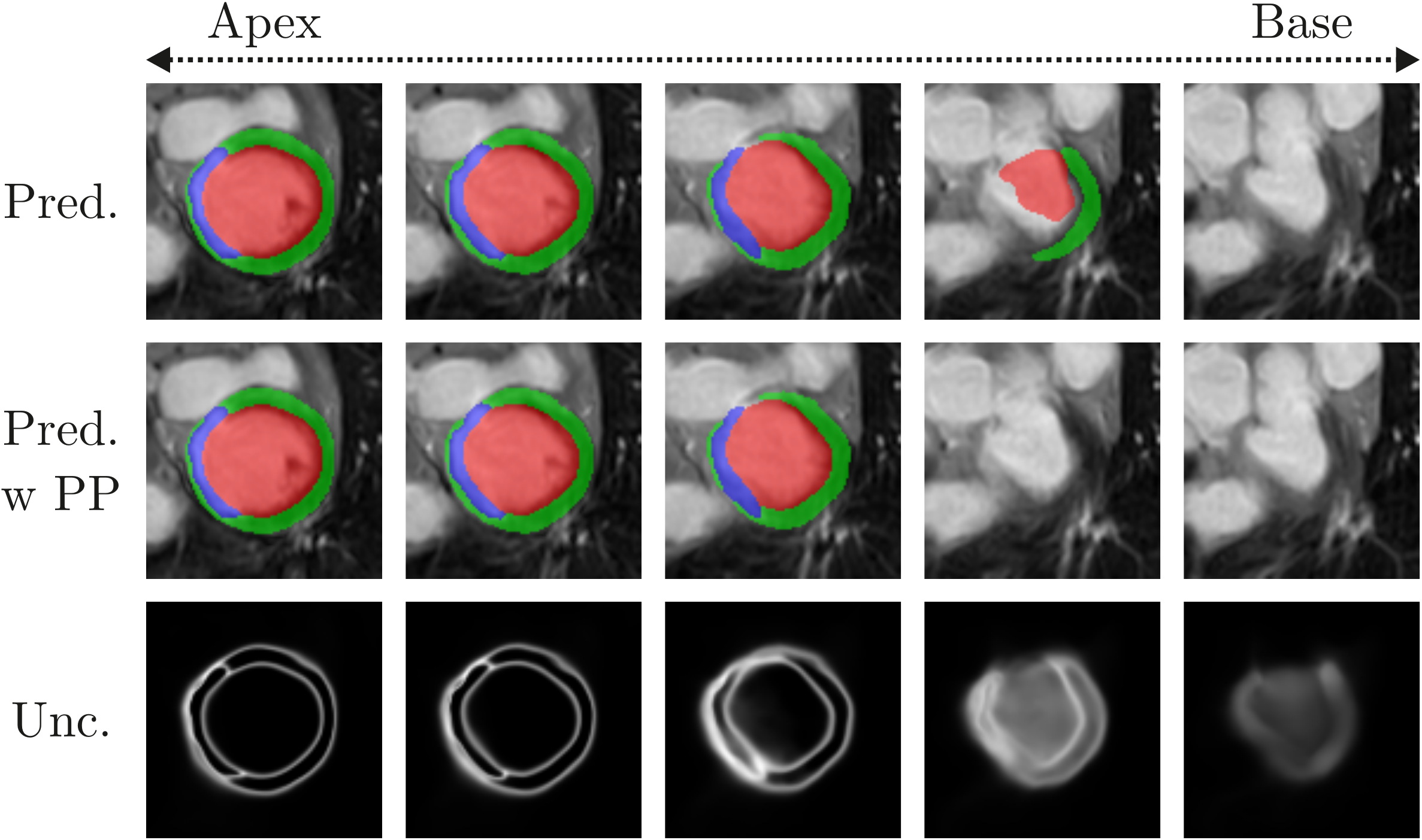}
\caption{
Predictions of \gls{carecnn} (row~1) are in some cases incomplete for the top-most slice towards the base of the left ventricle (col. 4). 
The model's uncertainty is computed as the entropy of the softmax prediction (row~3), where bright values indicate a higher uncertainty.
The highest uncertainty occurs in the incompletely labeled slice (col.~4).
This motivates our post-processing (PP) where, in this case, the incomplete prediction is removed (row~2).
}
\label{fig:uncertainty_topmost}
\end{figure}

\subsection{Inference}

As the subgroup (D8, M1, M12) for every image in the test set is known as well, we utilize the subgroup information for test set data to determine the final prediction as encouraged by the MYOSAIQ challenge organizers.
Specifically, we consider the label prediction $\labelpredstagethree$ of Stage~3 as the final label prediction $\labelpredfinal$ only for D8 data, while we use the Stage~2 label prediction $\labelpredstagetwo$ as the final label prediction $\labelpredfinal$ for M1 and M12 data.
The final label prediction $\labelpredfinal$ is defined as:
\begin{equation}
    \labelpredfinal =
    \begin{cases}
        \labelpredstagetwo &\text{if } \image \in \text{\{M1, M12\}}\\
        \labelpredstagethree &\text{if } \image \in \text{\{D8\}}.
    \end{cases}
\end{equation}
To further improve the final prediction of our method, we independently trained $N = 10$ \glspl{carecnn} with random weight initialization and random data augmentation.
These $N$ models were used as an ensemble for which the final label prediction is obtained by averaging the final label predictions of the individual models.
The average inference time per image for the whole ensemble with post-processing takes roughly 8~seconds using an NVIDIA GeForce RTX 3090.

\subsection{Post-Processing}

\begin{figure*}[t] 
\includegraphics[width=\textwidth]{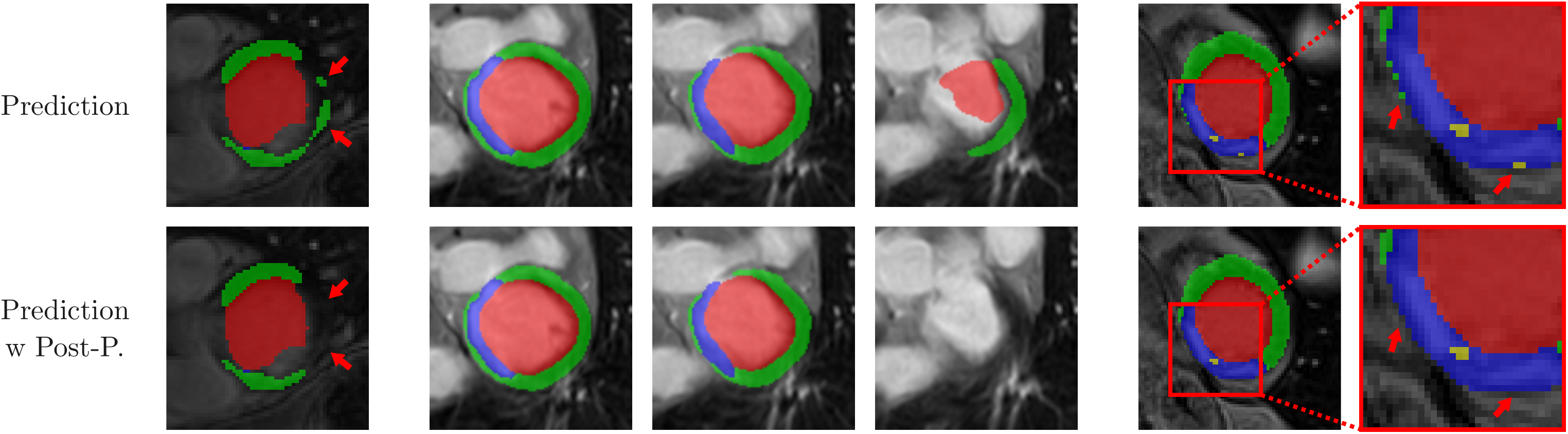}
\caption{
\gls{carecnn} predictions before (row 1) and after (row 2) post-processing.
Images refer to the proposed disconnected component removal (col. 1), the top-most slice removal (col. 2-4) and the outlier region replacement (col. 5-6).
Red arrows indicate regions of interest.
}
\label{fig:postprocessing}
\end{figure*}

As can be observed in Fig.~\ref{fig:uncertainty_topmost} (bottom row), after training on the data our \gls{carecnn} remains '\textit{uncertain}' about how far the heart should be segmented towards the base which may result in a top-most slice that is incompletely labeled.
Even though such incomplete model predictions in themselves are not incorrect, we decided to implement a series of post-processing steps to obtain more consistent predictions that take anatomical constraints into account.

As a first step of our post-processing pipeline, we employ a disconnected component removal strategy, where any components that are disconnected from the largest component in 3D as well as in-plane in 2D are removed.
In 3D, a connected component analysis is performed where all foreground labels are treated as one label and a 3D 6-connected kernel is applied.
Any independent region that is disconnected from the largest connected component is removed.
Due to the large slice thickness of the data, we also perform a connected component analysis for every in-plane 2D slice independently, following the same steps as described for the 3D variant and using a 2D 4-connected kernel in-plane.
The 2D strategy mostly affects the topmost slice that still contains foreground predictions and removes some smaller in-plane disconnected regions from that slice, see Fig.~\ref{fig:postprocessing} (col.~1).

Next, we propose a top-most slice removal strategy, where we compare the remaining foreground volume of the topmost slice that contains foreground predictions to the foreground volume of its neighboring slice towards the hearts' apex (i.e. the slice '\textit{below}' the top-most slice).
In case that the volume of the topmost slice is less than half the neighboring slice's volume, the topmost slice is removed completely.
An example is shown in Fig.~\ref{fig:postprocessing} (col.~4).

Lastly, an outlier region replacement strategy is applied, where very small regions of a single label are treated as outliers and are replaced if they are isolated from larger regions of the same label.
In the first step of this strategy, isolated regions are identified by performing a connected component analysis per label using a 3D 6-connected kernel.
Any region with a volume smaller than 0.1 ml is considered to be an outlier and undergoes a correction step, where the local neighborhood of each outlier voxel is observed to select a new label for that voxel, see Fig.~\ref{fig:postprocessing} (col.~5-6).
Specifically, we obtain label votes from all voxels within a 3D kernel of size $9 \times 9$ in-plane and $5$ out-of-plane, due to the large slice thickness.
This anisotropic kernel is sufficient as we perform a weighting based on a 3D Gaussian with sigma value 2 that considers the actual physical distance of any candidate.
Importantly, votes from voxels marked as outliers are not considered.
Finally, the maximum of the weighted votes indicates the most likely label for that voxel, which is then used as the label for that voxel.

\section{\uppercase{Experimental Setup}}

\subsection{Dataset}
\label{sec:dataset}

In this work, we used the publicly available dataset from the \gls{myosaiq} challenge\footnote{\url{https://www.creatis.insa-lyon.fr/Challenge/myosaiq/}, last accessed on October 8, 2023}, which was held in conjunction with \gls{fimh} 2023.
The aim of the \gls{myosaiq} challenge is to automatically segment four different cardiac structures from \gls{lge} \gls{mr} images of patients after myocardial infarction.
These structures encompass the \gls{lv}, \gls{myo}, \gls{mit} and \gls{mvo} if present.
The dataset consists of 467 \gls{lge} \gls{mr} images which are split into 376 training and 93 test images.
All images belong to one of three subgroups.
The first subgroup (D8) encompasses \gls{lge} images of 123 patients with acute myocardial infarction up to eight days after the infarction and originates from the MIMI-cohort~\cite{belle2016comparison}.
The second subgroup (M1) consists of \gls{lge} images of 204 patients, while the third subgroup (M12) contains \gls{lge} images of 140 patients, which were respectively obtained one and 12 months after coronary intervention and are part of the HIBISCUS-cohort.
For every image in the training dataset, a corresponding ground truth segmentation is available.
As the whole dataset consists of images after myocardial infarction, all ground truth segmentations in the dataset contain the \gls{lv}, \gls{myo} and \gls{mit} label.
However, the \gls{mvo} label is exclusive to the D8 subgroup and only present in roughly 66\% of the D8 data.
The in-plane physical resolution of the dataset varies from 0.9 to 2.2 mm and averages at 1.57 mm.
Out-of-plane, the physical resolution varies from 5 to 8 mm.

\subsection{Data Augmentation}

We augment training data using the training framework from \cite{payer2017multi,payer2019integrating} in 3D using random spatial and intensity transformations.
Spatially, we perform translation ($\pm 20$ voxels), rotation ($\pm 0.35$ radians), scaling (first isotropically with a factor between $[0.8, 1.2]$, then per dimension with a factor between $[0.9, 1.1]$) and elastic deformation (eight grid nodes per dimension, deformation values are sampled from $\pm 15$ voxels). 
For robust intensity normalization of the \gls{mr} images, the \nth{10} and \nth{90} percentile are linearly normalized to $-1$ and $1$, respectively. 
After normalization, a random intensity shift ($\pm 0.2$) as well as an intensity scaling with a factor between $[0.6, 1.4]$ is applied to the training image before modulating intensity values per label by an additional shift of ($\pm 0.2$) and scaling with a factor of $[0.9, 1.1]$.
All augmentation parameters are sampled uniformly from the respective value range.
Images of the test set are not augmented, however, they are robustly normalized identically to the training data to ensure similar intensity ranges.
To ensure consistency of the physical dimensions across the dataset, all training and test images are trilinearly resampled to an isotropic spacing of $1 \times 1 \times 1$ mm and an image size of $128 \times 128 \times 128$ voxel before being provided to the \gls{carecnn} model.

\subsection{Implementation Details}

At each stage of \gls{carecnn}, a U-Net-like~\cite{ronneberger2015u} network architecture is employed in 3D which follows the same structure, see Fig.~\ref{fig:overview}.
Similar to an encoder-decoder, the architecture can be separated into a contracting and an expanding path.
Importantly, by using skip-connections, the output of each level of the contracting path is concatenated to the input of the same level of the expanding path in the channel dimension.
At each of the five levels of the contracting and the expanding path, we use a single block consisting of two convolutions with an intermediate dropout layer~\cite{srivastava2014dropout}, after which a pooling or an upsampling layer is employed, respectively.
Two respectively three additional convolution layers are employed before and after the U-Net-like network of each stage.
All intermediate convolution layers use a $3 \times 3 \times 3$ kernel and $64$ filters, while the last convolution layer of each stage uses a $1 \times 1 \times 1$ kernel and as many filters as there are labels at the respective stage.
He initialization~\cite{he2015delving} is used to initialize all weights and the dropout rate is $0.1$.
We employ max pooling layers and tri-linear upsampling layers with a kernel size of $2 \times 2 \times 2$.
Leaky ReLU~\cite{maas2013rectifier} with a slope of $0.1$ is used after intermediate convolution layers, while a softmax activation is used after the last layer of each stage to compute the loss.
As optimizer, we employ Adam~\cite{kingma2014adam} with a learning rate of $0.001$, use an Exponential Moving Average strategy~\cite{laine2016temporal} with a decay of $0.999$ and train for $200,000$ iterations.
For each training iteration, we select one image with and one without the \gls{mvo} label which corresponds to a batch size of $2$ for the Stage~1 and Stage~2 models.
To ensure stable training, only images with the \gls{mvo} label are processed by the Stage~3 model, which results in an effective batch size of $1$ for that model.
During the development of our method, we trained our model only on $2/3$ of the training data and used the remaining $1/3$ of the data as a validation set.
For our submission to the challenge, we trained \gls{carecnn} on all training data and evaluation was performed on the hidden test set.
Final results were obtained by averaging the prediction of a \gls{carecnn} ensemble of $10$ models on the test set and $5$ models on the validation set.

\begin{table*}
\centering
\begin{tabular}{l | c c c c c c | c c c c}
\multicolumn{11}{c}{\cellcolor{lightgray} \textit{\tabbest{Mean over Labels}}} \\
\multirow{3}{*}{Team} & \multicolumn{2}{c}{DSC (\%)} & \multicolumn{2}{c}{HD (mm)} & \multicolumn{2}{c|}{ASSD (mm)} & \multirow{3}{*}{\shortstack[c]{CC\\ ($\uparrow$)}} & \multirow{3}{*}{\shortstack[c]{MAE\\ ($\downarrow$)}} & \multirow{3}{*}{\shortstack[c]{LOA\\ ($\downarrow$)}} & \multirow{3}{*}{\shortstack[c]{CRPS\\ ($\downarrow$)}} \\
& mean & std & mean & std & mean & std & & & & \\
& ($\uparrow$) & ($\downarrow$) & ($\downarrow$) & ($\downarrow$) & ($\downarrow$) & ($\downarrow$) & & & & \\
\cline{2-11}
\hline
gemr22 & 74.9 & 10.3 & \tabsec{13.452} & \tabsec{7.545} & \tabthi{0.711} & \tabsec{0.607} & 0.931 & 6.044 & 18.228 & \tabsec{0.011} \\
(proposed)$^{\dagger}$ & \tabbest{78.9} & \tabbest{8.5} & \tabbest{13.200} & 9.244 & \tabbest{0.574} & \tabbest{0.560} & \tabsec{0.938} & \tabbest{5.500} & \tabbest{16.140} & \tabbest{0.010} \\
akaroui & \tabthi{75.4} & 10.0 & 13.779 & 8.392 & \tabsec{0.697} & 0.689 & 0.929 & 5.827 & \tabthi{17.644} & 0.035 \\
\hline
Hairuiwang & \tabsec{75.6} & \tabthi{9.8} & 14.538 & 9.965 & \tabthi{0.711} & 0.673 & \tabthi{0.936} & \tabthi{5.810} & 18.219 & 0.038 \\
azanella & 75.1 & 10.6 & \tabthi{13.483} & \tabbest{7.195} & 0.724 & 0.685 & 0.905 & 5.842 & 18.337 & \tabbest{0.010} \\
KiwiYyy & 74.6 & 12.3 & 13.771 & \tabthi{7.626} & 0.734 & 0.702 & 0.930 & \tabsec{5.754} & \tabsec{17.247} & \tabbest{0.010} \\
hoanguyen93 & 74.3 & 11.0 & 13.905 & 9.009 & 0.744 & 0.758 & 0.904 & 5.924 & 19.297 & \tabbest{0.010} \\
nicoco & 73.7 & \tabsec{9.4} & 14.839 & 9.290 & 0.737 & \tabthi{0.630} & \tabsec{0.938} & 6.169 & 18.044 & 0.130 \\
hang\_jung & 73.7 & 9.9 & 15.063 & 8.752 & 0.835 & 0.766 & 0.907 & 6.415 & 19.802 & 0.014 \\
Dolphins & 73.4 & 11.9 & 15.711 & 10.061 & 0.754 & 0.675 & 0.911 & 6.578 & 20.556 & 0.042 \\
rrosales & 73.0 & 11.0 & 15.045 & 8.217 & 0.856 & 0.728 & \tabbest{0.940} & 6.788 & 19.442 & 0.020 \\
luiskabongo$^{\ddagger}$ & 72.3 & 10.6 & 15.584 & 9.561 & 0.804 & 0.718 & 0.917 & 7.099 & 21.622 & 0.105 \\
calderds & 72.0 & 11.7 & 17.321 & 11.853 & 0.849 & 0.767 & 0.909 & 6.628 & 20.820 & 0.039 \\
marwanabb & 69.8 & 12.9 & 15.667 & 9.499 & 1.131 & 1.183 & 0.883 & 7.534 & 22.477 & 0.071 \\
agaldran & 69.3 & 19.6 & 15.947 & 10.926 & - & - & 0.722 & 11.712 & 54.175 & - \\
Erwan & 65.5 & 12.3 & 20.502 & 8.416 & 1.200 & 1.022 & 0.853 & 7.204 & 22.358 & \tabthi{0.012} \\
farheenramzan & 55.3 & 10.6 & 20.594 & 9.051 & 1.641 & 1.233 & 0.720 & 9.349 & 26.389 & 0.016 \\
MYOSCANS & - & - & - & - & - & - & - & - & - & - \\
\hline
\end{tabular}
\caption{
Quantitative evaluation showing the mean score over all labels for ten metrics.
The proposed \gls{carecnn} is compared to the other \gls{myosaiq} challenge participants.
Invalid mean scores due to non-numeric results for at least one label are indicated by -.
The \tabbest{best}, \tabsec{second} and \tabthi{third} best scores are highlighted.
$^{\dagger}$Our teamname on the evaluation platform is 'ominous\_ocelot'.
$^{\ddagger}$Abbreviation for 'luiskabongo-inheart'.
}
\label{tab:mean_over_labels}
\end{table*}

\begin{table*}
\centering
\begin{tabular}{c | l | c c c c c c | c c c c}
\multicolumn{12}{c}{\cellcolor{lightgray} \textit{\tabbest{Best 3 Methods per Label}}} \\
& \multirow{3}{*}{Team} & \multicolumn{2}{c}{DSC (\%)} & \multicolumn{2}{c}{HD (mm)} & \multicolumn{2}{c|}{ASSD (mm)} & \multirow{3}{*}{\shortstack[c]{CC\\ ($\uparrow$)}} & \multirow{3}{*}{\shortstack[c]{MAE\\ ($\downarrow$)}} & \multirow{3}{*}{\shortstack[c]{LOA\\ ($\downarrow$)}} & \multirow{3}{*}{\shortstack[c]{CRPS\\ ($\downarrow$)}} \\
& & mean & std & mean & std & mean & std & & & & \\
& & ($\uparrow$) & ($\downarrow$) & ($\downarrow$) & ($\downarrow$) & ($\downarrow$) & ($\downarrow$) & & & & \\
\cline{3-12}
\hline
\multirow{4}{*}{\rotatebox[origin=c]{90}{LV}}
& \textbf{Overall Best} & \tabbest{93.7} & \tabbest{2.8} & \tabbest{6.406} & \tabbest{2.013} & \tabbest{0.392} & \tabbest{0.233} & \tabbest{0.980} & \tabbest{6.881} & \tabbest{17.121} & \tabbest{0.012} \\
\cline{2-12}
& gemr22 & 93.5 & 3.1 & 6.471 & 2.145 & 0.408 & 0.259 & \tabbest{0.980} & 7.308 & 18.533 & \tabbest{0.012} \\
& (proposed)$^{\dagger}$ & 93.4 & 3.4 & 6.666 & 2.155 & 0.419 & 0.290 & \tabbest{0.980} & \tabbest{6.881} & \tabbest{17.121} & \tabbest{0.012} \\
& akaroui & \tabbest{93.7} & 3.0 & \tabbest{6.406} & \tabbest{2.013} & \tabbest{0.392} & 0.264 & 0.978 & 7.313 & 18.768 & \tabbest{0.012} \\
\hline
\rowcolor{white} \multicolumn{12}{c}{} \\
\hline
\multirow{4}{*}{\rotatebox[origin=c]{90}{MYO}}
& \textbf{Overall Best} & \tabbest{82.2} & \tabbest{4.1} & \tabbest{11.753} & \tabbest{5.712} & \tabbest{0.390} & \tabbest{0.211} & \tabbest{0.967} & \tabbest{7.891} & \tabbest{22.251} & \tabbest{0.013} \\
\cline{2-12}
& gemr22 & 81.7 & 4.7 & 11.794 & 6.365 & 0.395 & 0.246 & 0.958 & 9.013 & 26.664 & 0.015 \\
& (proposed)$^{\dagger}$ & 81.6 & 5.0 & 12.839 & 7.144 & 0.405 & 0.253 & 0.954 & 9.845 & 25.686 & 0.016 \\
& akaroui & \tabbest{82.2} & 4.7 & 12.214 & 6.711 & \tabbest{0.390} & 0.259 & 0.964 & 8.463 & 24.263 & 0.014 \\
\hline
\rowcolor{white} \multicolumn{12}{c}{} \\
\hline
\multirow{4}{*}{\rotatebox[origin=c]{90}{MIT}}
& \textbf{Overall Best} & \tabbest{68.4} & \tabbest{16.1} & \tabbest{16.746} & \tabbest{12.482} & \tabbest{0.924} & \tabbest{1.310} & \tabbest{0.855} & \tabbest{4.044} & \tabbest{17.866} & \tabbest{0.007} \\
\cline{2-12}
& gemr22 & 66.0 & 17.1 & 18.201 & \tabbest{12.482} & 1.005 & 1.431 & 0.799 & 4.510 & 20.197 & 0.008 \\
& (proposed)$^{\dagger}$ & \tabbest{68.4} & \tabbest{16.1} & \tabbest{16.746} & 13.414 & \tabbest{0.924} & 1.377 & 0.833 & \tabbest{4.044} & 18.647 & \tabbest{0.007} \\
& akaroui & 65.8 & 17.4 & 19.790 & 14.751 & 1.092 & 1.666 & 0.789 & 4.582 & 20.873 & 0.008 \\
\hline
\rowcolor{white} \multicolumn{12}{c}{} \\
\hline
\multirow{4}{*}{\rotatebox[origin=c]{90}{MVO}}
& \textbf{Overall Best} & \tabbest{72.0} & \tabbest{9.5} & \tabbest{14.539} & \tabbest{5.682} & \tabbest{0.547} & \tabbest{0.321} & \tabbest{0.995} & \tabbest{1.231} & \tabbest{3.106} & \tabbest{0.003} \\
\cline{2-12}
& gemr22 & 58.5 & 16.4 & 17.343 & 9.187 & 1.037 & 0.492 & 0.987 & 3.343 & 7.516 & 0.008 \\
& (proposed)$^{\dagger}$ & \tabbest{72.0} & \tabbest{9.5} & 16.548 & 14.261 & \tabbest{0.547} & \tabbest{0.321} & 0.985 & \tabbest{1.231} & \tabbest{3.106} & \tabbest{0.003} \\
& akaroui & 59.9 & 15.0 & 16.705 & 10.092 & 0.913 & 0.566 & 0.984 & 2.950 & 6.671 & 0.106 \\
\hline
\end{tabular}
\caption{
Quantitative evaluation showing the individual label scores of the three best \gls{myosaiq} challenge participants for ten metrics.
'Overall Best' refers to the best score obtained by any participant and is used as an upper baseline for each label and metric.
The $\tabbest{best}$ score for each metric considering all 18 participants is highlighted in bold.
$^{\dagger}$Our teamname on the evaluation platform is 'ominous\_ocelot'.
}
\label{tab:best_three_per_label}
\end{table*}

\begin{table*}
\centering
\begin{tabular}{c | c | c c c c c c | c c c c}
\multicolumn{12}{c}{\cellcolor{lightgray} \textit{\textbf{Ablation of Proposed CaRe-CNN Ensemble}}} \\
\multirow{3}{*}{PP} & \multirow{3}{*}{Label} & \multicolumn{2}{c}{DSC (\%)} & \multicolumn{2}{c}{HD (mm)} & \multicolumn{2}{c|}{ASSD (mm)} & \multirow{3}{*}{\shortstack[c]{CC\\ ($\uparrow$)}} & \multirow{3}{*}{\shortstack[c]{MAE\\ ($\downarrow$)}} & \multirow{3}{*}{\shortstack[c]{LOA\\ ($\downarrow$)}} & \multirow{3}{*}{\shortstack[c]{CRPS\\ ($\downarrow$)}} \\
& & mean & std & mean & std & mean & std & & & & \\
& & ($\uparrow$) & ($\downarrow$) & ($\downarrow$) & ($\downarrow$) & ($\downarrow$) & ($\downarrow$) & & & & \\
\cline{3-12}
\hline
\multirow{4}{*}{\checkx}
& LV & 93.4 & 3.3 & 6.892 & 2.157 & 0.422 & 0.294 & 0.980 & 6.837 & 17.215 & 0.011 \\
& MYO & 81.6 & 4.8 & 12.088 & 6.611 & 0.400 & 0.238 & 0.957 & 9.944 & 25.271 & 0.016 \\
& MIT & 68.5 & 15.9 & 16.892 & 13.46 & 0.901 & 1.346 & 0.837 & 4.000 & 18.491 & 0.007 \\
& MVO & 71.7 & 10.0 & 17.569 & 13.853 & 0.576 & 0.329 & 0.985 & 1.210 & 3.104 & 0.003 \\
\hline
\multirow{4}{*}{\checkmark}
& LV & 93.4 & 3.4 & 6.666 & 2.155 & 0.419 & 0.290 & 0.980 & 6.881 & 17.121 & 0.012 \\
& MYO & 81.6 & 5.0 & 12.839 & 7.144 & 0.405 & 0.253 & 0.954 & 9.845 & 25.686 & 0.016 \\
& MIT & 68.4 & 16.1 & 16.746 & 13.414 & 0.924 & 1.377 & 0.833 & 4.044 & 18.647 & 0.007 \\
& MVO & 72.0 & 9.5 & 16.548 & 14.261 & 0.547 & 0.321 & 0.985 & 1.231 & 3.106 & 0.003 \\
\hline
\multicolumn{2}{c|}{\textbf{Mean Diff.}} & \tabbetter{+0.1} & \textbf{0} & \tabbetter{-0.161} & \tabworse{+0.223} & \tabbetter{-0.001} & \tabworse{+0.009} & \tabworse{-0.002} & \tabworse{+0.003} & \tabworse{+0.120} & \tabworse{+0.000} \\
\hline
\end{tabular}
\caption{
Ablation of the proposed post-processing (PP) when applied to our \gls{carecnn} ensemble predictions.
Scores before~(\checkx) and after~(\checkmark) post-processing are shown for each label and ten metrics.
The last row refers to the mean difference, where improvements when using post-processing are highlighted in $\tabbetter{green}$, while declines are highlighted in $\tabworse{red}$.
}
\label{tab:ablation}
\end{table*}

\begin{figure*}[t] 
\includegraphics[width=\textwidth]{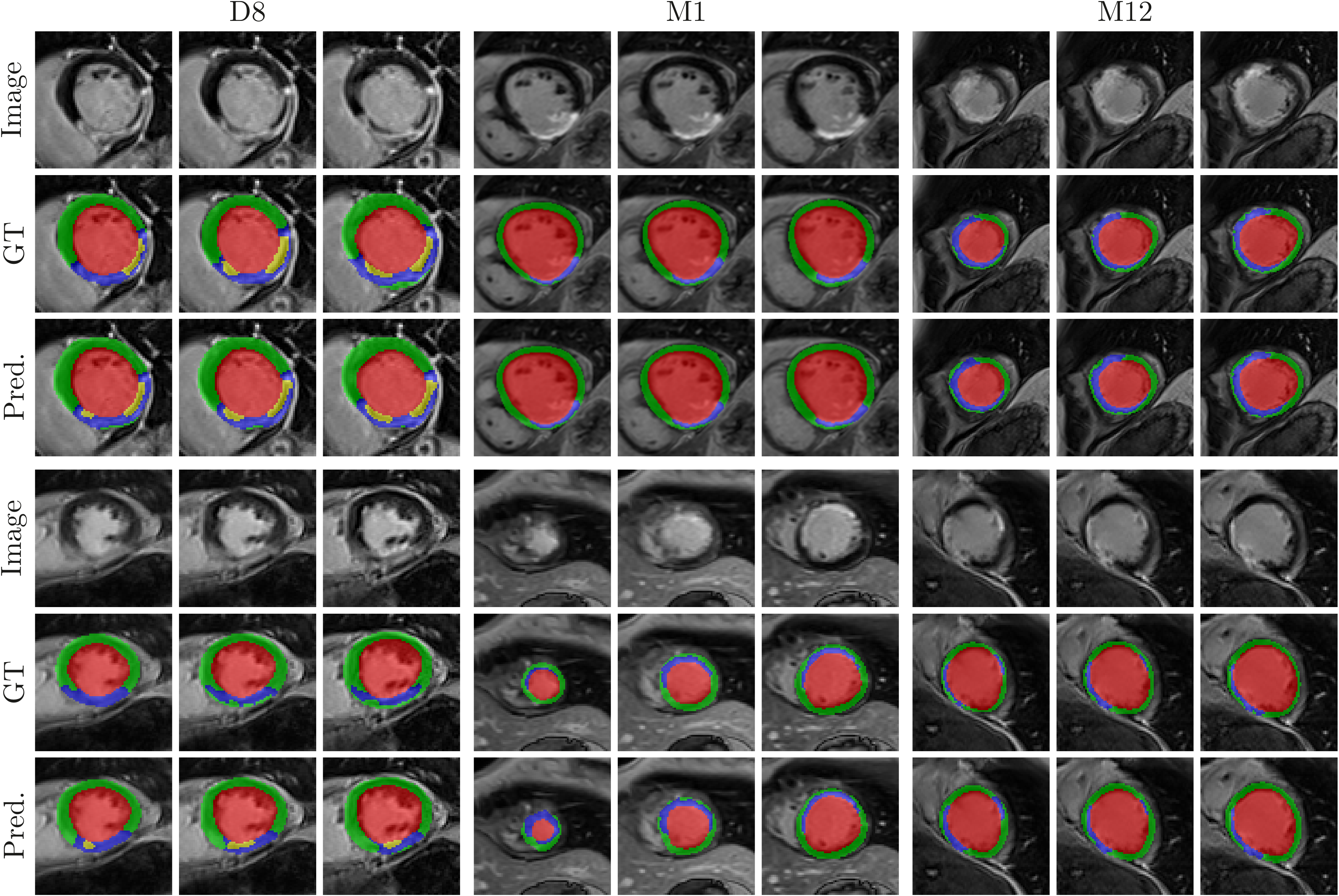}
\caption{
Qualitative results of \gls{carecnn} on the validation set.
Columns refer to three consecutive slices of \gls{lge} \gls{mr} scans of patients after \gls{mi} for the three subgroups: D8 (col. 1-3), M1 (col. 4-6) and M12 (7-9).
Rows refer to scans of two separate patients and show the image (rows 1, 4), ground truth (rows 2, 5) and prediction of \gls{carecnn} (rows 3, 6).
}
\label{fig:qualitative_results_valset}
\end{figure*}

\begin{figure*}[t] 
\includegraphics[width=\textwidth]{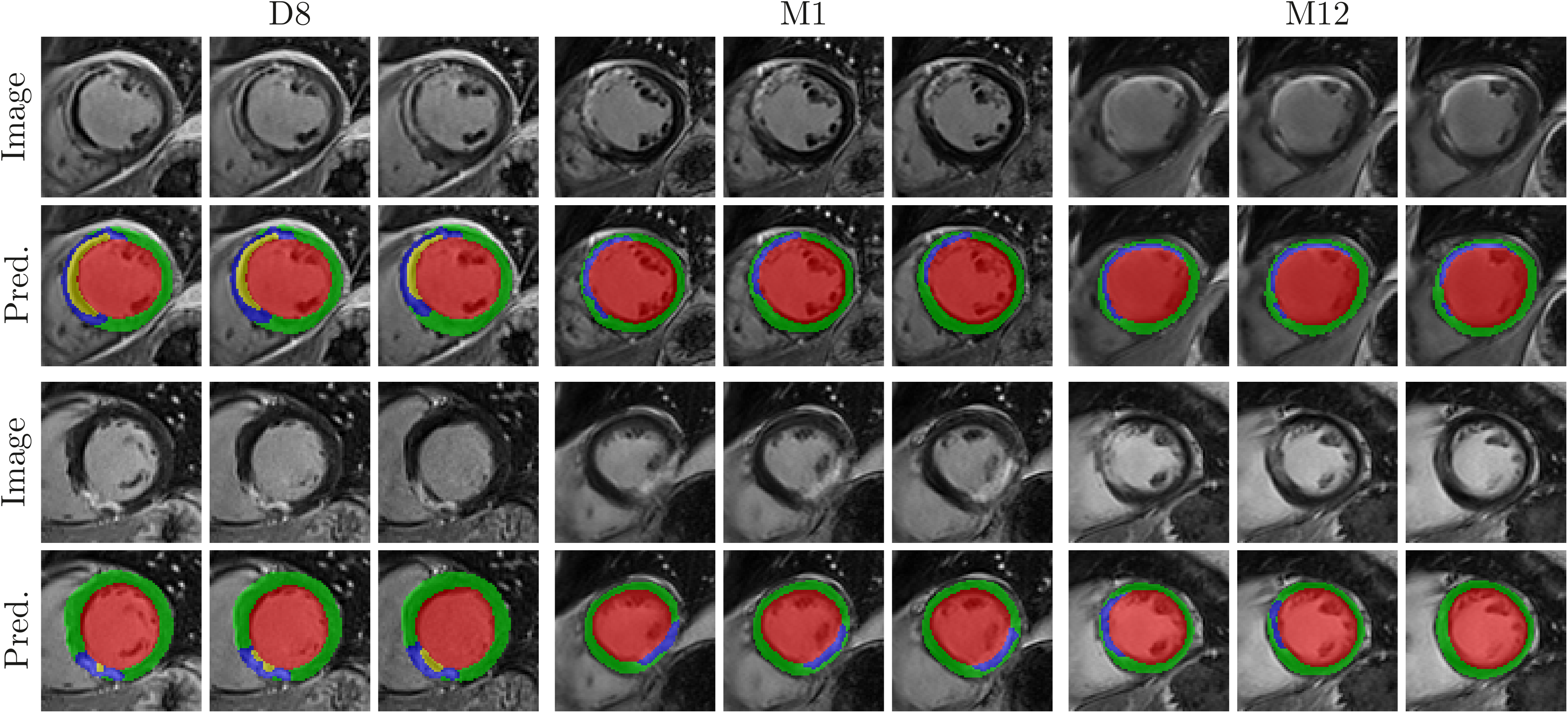}
\caption{
Qualitative results of \gls{carecnn} on the test set.
Columns refer to three consecutive slices of \gls{lge} \gls{mr} scans of patients after \gls{mi} for each subgroup: D8 (col. 1-3), M1 (col. 4-6) and M12 (7-9).
Rows refer to scans of two separate patients and show the image (rows 1, 3) and prediction of \gls{carecnn} (rows 2, 4).
Ground truth is not available for the test set.
}
\label{fig:qualitative_results_testset}
\end{figure*}

\section{\uppercase{Results}}

The quantitative evaluation is performed by comparing our \gls{carecnn} method to the other 17 participants of the \gls{myosaiq} challenge on the hidden test set.
For each participant, we obtained ten metric scores for each label individually from the official evaluation platform\footnote{\url{https://codalab.lisn.upsaclay.fr/competitions/13631}, last accessed on October 8, 2023}, which is publicly available.
The used metrics respectively encompass the mean and standard deviation of the \gls{dsc} in percent as well as the \gls{hd} and \gls{assd} in mm.
Furthermore, the list of metrics includes the mean \gls{cc}, \gls{mae}, \gls{loa} and the \gls{crps}.

In order to summarize the results, we computed the mean score over the four labels for each metric and present them in Table~\ref{tab:mean_over_labels} for each participant.
This is also true for the standard deviation of \gls{dsc}, \gls{hd} and \gls{assd}, where we also computed the mean score over the labels.
The best score for each metric is given in bold, while the second and third best metric scores are shown in underlined blue and italicized orange, respectively.

Table~\ref{tab:best_three_per_label} presents quantitative results per label for each metric to give some insight into the individual scores.
In the interest of space, we only provide the scores for the three best performing methods in the challenge as announced at the \gls{fimh} 2023 conference.
Nevertheless, to indicate the overall best score over all teams for each metric and label, we additionally show the best score obtained by \textit{any} participant as an upper bound baseline.
The best score for each metric when considering all 18 challenge participants is given in bold.

Table~\ref{tab:ablation} shows an ablation of the proposed \gls{carecnn} ensemble with and without post-processing.
Again, the scores were obtained from the evaluation platform of the challenge, where we submitted our prediction results from the exact same models with and without post-processing.
We show the score for each label and all metrics evaluated in the challenge.
The last row represents the mean difference between the scores obtained with and without post-processing.
Underlined green numbers indicate an improvement and red numbers refer to a decline in performance when post-processing is applied compared to when it is not.

The qualitative evaluation of our \gls{carecnn} is performed by visually inspecting the predictions.
As ground truth segmentations for the test set data are hidden, we also present qualitative results of \gls{carecnn} trained on $2/3$ and validated on $1/3$ of the actual training data for the \gls{myosaiq} challenge in Fig.~\ref{fig:qualitative_results_valset} to allow a comparison of our predictions to the ground truth.
Additionally, we provide qualitative results of our final method submitted to the challenge on the test set in Fig.~\ref{fig:qualitative_results_testset}, however, without publicly available ground truth segmentations, the predictions are only compared to the respective input images.
Both figures show three consecutive slices of two \gls{mr} scans of patients after acute \gls{mi} per subgroup (D8, M1, M12).

\section{\uppercase{Discussion}}

\paragraph{Quantitative Evaluation}

The mean score over the four labels presented in Table~\ref{tab:mean_over_labels} shows, that on average our method achieved the best score for eight out of ten metrics.
Other participants only outperformed our method on the mean standard deviation of the \gls{hd} as well as the \gls{cc}, where \gls{carecnn} obtained a tied second best score.
Most notably, our method shows great improvements compared to the other methods on the \gls{dsc} and \gls{assd} scores.
Specifically, with a mean \gls{dsc} of $78.9\%$, \gls{carecnn} achieved 
an improvement of $3.3\%$ compared to the second best method with $75.6\%$.
The same $3.3\%$ window applied to the range $[75.6\%, 72.3\%]$ encompasses the second up to the \nth{12} best mean \gls{dsc} score.
Similarly, with a result of $0.574 \text{ mm}$ on the \gls{assd} score our method achieved an improvement of $0.123 \text{ mm}$ over the second best \gls{assd} score with $0.697 \text{ mm}$.
The second up to the \nth{10} best score lie within the same $0.123 \text{ mm}$ window of $[0.697 \text{ mm}, 0.820 \text{ mm}]$.

More details are provided in Table~\ref{tab:best_three_per_label}, where the per label scores and the overall best score of \textit{any} method are shown.
For the \gls{lv} results, it can be observed that our method obtained the best scores for \gls{mae} and \gls{loa}, and obtained tied best scores with other methods for the \gls{cc} and \gls{crps}.
Moreover, the shown \gls{dsc}, \gls{hd} and \gls{assd} scores are all very close to one another with our method achieving $93.4\%$ (best: $93.7\%$) \gls{dsc}, $6.67 \text{ mm}$ (best: $6.41 \text{ mm}$) \gls{hd} and $0.42 \text{ mm}$ (best: $0.39 \text{ mm}$) \gls{assd}.
On \gls{myo}, our method did not obtain the best score on any metric and underperformed compared to the overall best score most notably with $12.839 \text{ mm}$ (best: $11.753 \text{ mm}$) \gls{hd}, $9.845$ (best: $7.891$) \gls{mae} and $25.686$ (best: $22.251$) \gls{loa}.
Nevertheless, on other metrics like \gls{dsc} and \gls{assd} our method remains competitive to the other methods achieving $81.6\%$ (best: $82.2\%$) \gls{dsc} and $0.405 \text{ mm}$ (best: $0.390 \text{ mm}$) \gls{assd}.

Compared to the other challenge participants, our \gls{carecnn} excelled when segmenting the difficult but clinically most relevant \gls{mit} and \gls{mvo} labels, where our method obtained the best score for six and seven out of the ten metrics, respectively.
Among the three challenge winners, our method achieved good improvements on the \gls{mit} label with $68.4\%$ ($+2.4\%$) \gls{dsc}, $16.746 \text{ mm}$ ($-1.455 \text{ mm}$) \gls{hd}, $0.924 \text{ mm}$ ($-0.081 \text{ mm}$) \gls{assd} and $4.044$ ($-0.466$) \gls{mae}.
Interestingly, our method underperformed on the \gls{hd} of the \gls{mvo} label achieving a mean of $16.548 \text{ mm}$ (best: $14.539 \text{ mm}$) and a standard deviation of $14.261 \text{ mm}$ (best: $5.682 \text{ mm}$).
On other metrics, however, \gls{carecnn} achieved great improvements for the \gls{mvo} label compared to the other two challenge winners, namely $72.0\%$ ($+12.1\%$) \gls{dsc}, $0.547 \text{ mm}$ ($-0.366 \text{ mm}$) \gls{assd}, $1.231$ ($-1.719$) \gls{mae} and $3.106$ ($-3.565$) \gls{loa}.

\paragraph{Post-processing}

When observing the training data more closely, we noticed that the ground truth annotations of the heart labels towards the base of the heart are not always complete.
Most notably, how far slices are labeled towards the base varies from image to image, which is likely an artifact from the annotation protocol.
While such incomplete annotations are not incorrect, they introduce a bias to the dataset which is reflected by a machine learning model and leads to some expected inconsistencies in the model predictions.
We mitigate these inconsistencies using a series of post-processing steps to obtain more consistent predictions and show that quantitative scores for all metrics are almost unchanged in Table~\ref{tab:ablation}.
The most affected metric is the \gls{hd} resulting in a mean of $13.200 \text{ mm}$ (mean difference: $-0.161 \text{ mm}$) and a standard deviation of $9.243 \text{ mm}$ (mean difference: $+0.223 \text{ mm}$) after post-processing.
This confirms our expectation, that the top-most slice removal strategy paired with the large slice thickness of $5.6 \text{ mm}$ on average leads to the \gls{hd} being the most affected metric as it is defined as the maximum distance of \textit{any} voxel-pair of the same label between ground truth and prediction.
Nevertheless, the relative change over all metrics averages to $0.6\%$ when using post-processing, which confirms that it can be safely applied in order to improve the qualitative consistency of the predictions.

\paragraph{Qualitative Evaluation}

The qualitative results on the validation set in Fig.~\ref{fig:qualitative_results_valset} confirm that most label predictions are very close to the ground truth.
On closer inspection, however, some differences can be spotted.
For example, one of the two \gls{mvo} regions is predicted in one additional consecutive slice in contrast to the ground truth (D8, top), while the \gls{mit} label is overpredicted close to the apex (M1, bottom).
Also, an \gls{mvo} label prediction for a patient without \gls{mvo} is visible (D8, bottom).
Nevertheless, many regions are predicted correctly, most notably even for data where the wall is in parts only two to three voxels thick (M12, bottom). 
On the test set in Fig.~\ref{fig:qualitative_results_valset}, qualitative results can only be compared to the \gls{lge} \gls{mr} image.
Overall, the label predictions appear to be realistic which is supported by our quantitative evaluation, however, further confirmation needs to be performed by an expert.

\paragraph{Challenges and Limitations}

One major challenge of correctly segmenting the structures of interest arises from the limited resolution of the \gls{lge} \gls{mr} data in combination with the shape and small physical size of the structures, most notably the \gls{mit} and \gls{mvo} label.
While the \gls{lv} is comparatively easy to segment due to its size and blob-like shape in 3D, the \gls{fmyo} label that surrounds the \gls{lv} averages to a mid-diastolic thickness of $6.47 \pm 1.07 \text{ mm}$ in women and $7.90 \pm 1.24 \text{ mm}$ in men~\cite{walpot2019left} without considering infarction.
In a small cohort, \cite{khalid2019assessing} showed that during ejection, healthy wall segments are roughly three times as thick ($8.73 \text{ mm}$) compared with infarcted wall segments ($2.86 \text{ mm}$).
Furthermore, infarction might only affect some part of the myocardial tissue in transmural direction such that two or even all three of the \gls{fmyo} sublabels (\gls{myo}, \gls{mit} and \gls{mvo}) might be present across the already thin wall.
The in-plane resolution of the \gls{lge} \gls{mr} data with $1.57 \text{ mm}$ on average paired with the small physical size of some of the structures of interest leads to a potential transmural thickness of only a few voxels for these labels.
Moreover, segmentation models are inherently uncertain near the label borders and thus, prone to single voxel errors, which can strongly affect the scores for small structures like the \gls{mit} and \gls{mvo} labels.
The combination of these effects explains the disparity of the \gls{lv} to the \gls{mit} and \gls{mvo} label scores for which \gls{carecnn} achieved the best score in six (\gls{mit}) and seven (\gls{mvo}) out of ten metrics among 18 challenge participants.

A remaining challenge arises from the \gls{mvo} label predictions for some patients of the D8 subgroup, where the label was not predicted when it should be present or vice versa.
Since the presence of \gls{mvo} is linked to an increased risk of adverse cardiovascular events~\cite{hamirani2014effect,rios2019microvascular}, incorrect predictions of \gls{mvo} might impact clinical decision making if trusted blindly.
While manual verification by an expert is necessary, our state-of-the-art predictions can alleviate the manual workload to obtain correct segmentations of patient-specific anatomy.

\section{\uppercase{Conclusion}}

In this work we presented \gls{carecnn}, a 3-stage cascading refinement \gls{cnn}, which segments cardiac \gls{lge} \gls{mr} images after \gls{mi}.
The cascading architecture is designed to exploit the hierarchical label definition of the data and is trained end-to-end fully in 3D.
Furthermore, we employed a series of post-processing steps that improve the consistency of the predictions by taking anatomical constraints into account.
The proposed \gls{carecnn} was submitted to the \gls{myosaiq} challenge, where it ranked second out of 18 participating teams and achieved state-of-the-art segmentation results, most notably when segmenting the difficult \gls{mit} and \gls{mvo} labels.
Due to great improvements over related work on the difficult but clinically very relevant \gls{mvo} label, our method obtained the best score in eight out of ten metrics when computing the mean over all labels.
Precise segmentations of healthy and infarcted myocardial tissue after \gls{mi} allow patient-specific therapy planning and are an important step towards personalized medicine.
In our future work, we plan to investigate uncertainty quantification strategies to further improve \gls{carecnn} for future rounds of the \gls{myosaiq} challenge.

\section*{\uppercase{Acknowledgements}}

This research was funded by the InstaTwin grant FO999891133 from the Austrian Research Promotion Agency (FFG).

\bibliographystyle{apalike}
{\small
\bibliography{refs}}

\begin{thebibliography}{}

\bibitem[Akkus et~al., 2017]{akkus2017deep}
Akkus, Z., Galimzianova, A., Hoogi, A., Rubin, D.~L., and Erickson, B.~J. (2017).
\newblock {Deep Learning for Brain MRI Segmentation: State of the Art and Future Directions}.
\newblock {\em Journal of Digital Imaging}, 30:449--459.

\bibitem[Belle et~al., 2016]{belle2016comparison}
Belle, L., Motreff, P., Mangin, L., Rang{\'e}, G., Marcaggi, X., Marie, A., Ferrier, N., Dubreuil, O., Zemour, G., Souteyrand, G., et~al. (2016).
\newblock {Comparison of Immediate with Delayed Stenting Using the Minimalist Immediate Mechanical Intervention Approach in Acute ST-Segment--Elevation Myocardial Infarction: The MIMI Study}.
\newblock {\em Circulation: Cardiovascular Interventions}, 9(3):e003388.

\bibitem[Campos et~al., 2022]{Campos2022}
Campos, F.~O., Neic, A., {Mendonca Costa}, C., Whitaker, J., O'Neill, M., Razavi, R., Rinaldi, C.~A., Scherr, D., Niederer, S.~A., Plank, G., et~al. (2022).
\newblock {An Automated Near-Real Time Computational Method for Induction and Treatment of Scar-related Ventricular Tachycardias.}
\newblock {\em Medical Image Analysis}, 80:102483.

\bibitem[Chen et~al., 2020]{chen2020deep}
Chen, C., Qin, C., Qiu, H., Tarroni, G., Duan, J., Bai, W., and Rueckert, D. (2020).
\newblock {Deep Learning for Cardiac Image Segmentation: A Review}.
\newblock {\em Frontiers in Cardiovascular Medicine}, 7:25.

\bibitem[Chen et~al., 2022]{chen2022automatic}
Chen, Z., Lalande, A., Salomon, M., Decourselle, T., Pommier, T., Qayyum, A., Shi, J., Perrot, G., and Couturier, R. (2022).
\newblock {Automatic Deep Learning-based Myocardial Infarction Segmentation from Delayed Enhancement MRI}.
\newblock {\em Computerized Medical Imaging and Graphics}, 95:102014.

\bibitem[Esteva et~al., 2017]{Esteva2017}
Esteva, A., Kuprel, B., Novoa, R.~A., Ko, J., Swetter, S.~M., Blau, H.~M., and Thrun, S. (2017).
\newblock {Dermatologist-level Classification of Skin Cancer with Deep Neural Networks}.
\newblock {\em Nature}, 542(7639):115--118.

\bibitem[Fahmy et~al., 2018]{fahmy2018automated}
Fahmy, A.~S., Rausch, J., Neisius, U., Chan, R.~H., Maron, M.~S., Appelbaum, E., Menze, B., and Nezafat, R. (2018).
\newblock {Automated Cardiac MR Scar Quantification in Hypertrophic Cardiomyopathy Using Deep Convolutional Neural Networks}.
\newblock {\em JACC: Cardiovascular Imaging}, 11(12):1917--1918.

\bibitem[Feng et~al., 2022]{Feng2022-md}
Feng, S., Liu, Q., Patel, A., Bazai, S.~U., Jin, C.-K., Kim, J.~S., Sarrafzadeh, M., Azzollini, D., Yeoh, J., Kim, E., et~al. (2022).
\newblock {Automated Pneumothorax Triaging in Chest X-rays in the New Zealand Population Using Deep-learning Algorithms}.
\newblock {\em Journal of Medical Imaging and Radiation Oncology}, 66(8):1035--1043.

\bibitem[Gillette et~al., 2021]{gillette2021framework}
Gillette, K., Gsell, M.~A., Prassl, A.~J., Karabelas, E., Reiter, U., Reiter, G., Grandits, T., Payer, C., {\v{S}}tern, D., Urschler, M., et~al. (2021).
\newblock {A Framework for the Generation of Digital Twins of Cardiac Electrophysiology from Clinical 12-leads ECGs}.
\newblock {\em Medical Image Analysis}, 71:102080.

\bibitem[Graves et~al., 2013]{graves2013speech}
Graves, A., Mohamed, A.-r., and Hinton, G. (2013).
\newblock {Speech Recognition with Deep Recurrent Neural Networks}.
\newblock In {\em Proceedings of the IEEE International Conference on Acoustics, Speech and Signal Processing}, pages 6645--6649.

\bibitem[Hamirani et~al., 2014]{hamirani2014effect}
Hamirani, Y.~S., Wong, A., Kramer, C.~M., and Salerno, M. (2014).
\newblock {Effect of Microvascular Obstruction and Intramyocardial Hemorrhage by CMR on LV Remodeling and Outcomes after Myocardial Infarction: A Systematic Review and Meta-Analysis}.
\newblock {\em JACC: Cardiovascular Imaging}, 7(9):940--952.

\bibitem[He et~al., 2015]{he2015delving}
He, K., Zhang, X., Ren, S., and Sun, J. (2015).
\newblock {Delving Deep into Rectifiers: Surpassing Human-Level Performance on ImageNet Classification}.
\newblock In {\em Proceedings of the IEEE International Conference on Computer Vision}, pages 1026--1034.

\bibitem[Hellermann et~al., 2002]{hellermann2002heart}
Hellermann, J.~P., Jacobsen, S.~J., Gersh, B.~J., Rodeheffer, R.~J., Reeder, G.~S., and Roger, V.~L. (2002).
\newblock {Heart Failure after Myocardial Infarction: A Review}.
\newblock {\em The American Journal of Medicine}, 113(4):324--330.

\bibitem[Khalid et~al., 2019]{khalid2019assessing}
Khalid, A., Lim, E., Chan, B.~T., Abdul~Aziz, Y.~F., Chee, K.~H., Yap, H.~J., and Liew, Y.~M. (2019).
\newblock {Assessing Regional Left Ventricular Thickening Dysfunction and Dyssynchrony via Personalized Modeling and 3D Wall Thickness Measurements for Acute Myocardial Infarction}.
\newblock {\em Journal of Magnetic Resonance Imaging}, 49(4):1006--1019.

\bibitem[Kim et~al., 1999]{kim1999relationship}
Kim, R.~J., Fieno, D.~S., Parrish, T.~B., Harris, K., Chen, E.-L., Simonetti, O., Bundy, J., Finn, J.~P., Klocke, F.~J., and Judd, R.~M. (1999).
\newblock {Relationship of MRI Delayed Contrast Enhancement to Irreversible Injury, Infarct Age, and Contractile Function}.
\newblock {\em Circulation}, 100(19):1992--2002.

\bibitem[Kim and Manning, 2004]{kim2004viability}
Kim, R.~J. and Manning, W.~J. (2004).
\newblock {Viability Assessment by Delayed Enhancement Cardiovascular Magnetic Resonance: Will Low-dose Dobutamine Dull the Shine?}
\newblock {\em Circulation}, 109(21):2476--2479.

\bibitem[Kingma and Ba, 2015]{kingma2014adam}
Kingma, D.~P. and Ba, J.~L. (2015).
\newblock {Adam: A Method for Stochastic Optimization}.
\newblock In {\em Proceedings of the International Conference on Learning Representations}.

\bibitem[Laine and Aila, 2016]{laine2016temporal}
Laine, S. and Aila, T. (2016).
\newblock {Temporal Ensembling for Semi-Supervised Learning}.
\newblock In {\em Proceedings of the International Conference on Learning Representations}.

\bibitem[Lalande et~al., 2022]{lalande2022deep}
Lalande, A., Chen, Z., Pommier, T., Decourselle, T., Qayyum, A., Salomon, M., Ginhac, D., Skandarani, Y., Boucher, A., Brahim, K., et~al. (2022).
\newblock {Deep Learning Methods for Automatic Evaluation of Delayed Enhancement-MRI. The Results of the EMIDEC Challenge}.
\newblock {\em Medical Image Analysis}, 79:102428.

\bibitem[Maas et~al., 2013]{maas2013rectifier}
Maas, A.~L., Hannun, A.~Y., and Ng, A.~Y. (2013).
\newblock {Rectifier Nonlinearities Improve Neural Network Acoustic Models}.
\newblock In {\em Proceedings of the International Conference on Machine Learning}, volume~30, page~3. Atlanta, GA.

\bibitem[Moccia et~al., 2019]{moccia2019development}
Moccia, S., Banali, R., Martini, C., Muscogiuri, G., Pontone, G., Pepi, M., and Caiani, E.~G. (2019).
\newblock {Development and Testing of a Deep Learning-based Strategy for Scar Segmentation on CMR-LGE Images}.
\newblock {\em Magnetic Resonance Materials in Physics, Biology and Medicine}, 32:187--195.

\bibitem[Payer et~al., 2017]{payer2017multi}
Payer, C., {\v{S}}tern, D., Bischof, H., and Urschler, M. (2017).
\newblock {Multi-label Whole Heart Segmentation using CNNs and Anatomical Label Configurations}.
\newblock In {\em International Workshop on Statistical Atlases and Computational Models of the Heart}, pages 190--198. Springer.

\bibitem[Payer et~al., 2019]{payer2019integrating}
Payer, C., {\v{S}}tern, D., Bischof, H., and Urschler, M. (2019).
\newblock {Integrating Spatial Configuration into Heatmap Regression based CNNs for Landmark Localization}.
\newblock {\em Medical Image Analysis}, 54:207--219.

\bibitem[Payer et~al., 2020]{Payer2020-yv}
Payer, C., {\v{S}}tern, D., Bischof, H., and Urschler, M. (2020).
\newblock {Coarse to Fine Vertebrae Localization and Segmentation with SpatialConfiguration-Net and U-Net}.
\newblock In {\em 15th International Joint Conference on Computer Vision, Imaging and Computer Graphics Theory and Applications ({VISIGRAPP} 2020) - Volume 5: {VISAPP}}, pages 124--133.

\bibitem[Perin et~al., 2002]{perin2002assessing}
Perin, E.~C., Silva, G.~V., Sarmento-Leite, R., Sousa, A.~L., Howell, M., Muthupillai, R., Lambert, B., Vaughn, W.~K., and Flamm, S.~D. (2002).
\newblock {Assessing Myocardial Viability and Infarct Transmurality with Left Ventricular Electromechanical Mapping in Patients with Stable Coronary Artery Disease: Validation by Delayed-Enhancement Magnetic Resonance Imaging}.
\newblock {\em Circulation}, 106(8):957--961.

\bibitem[Rios-Navarro et~al., 2019]{rios2019microvascular}
Rios-Navarro, C., Marcos-Garces, V., Bayes-Genis, A., Husser, O., Nunez, J., and Bodi, V. (2019).
\newblock {Microvascular Obstruction in ST-segment Elevation Myocardial Infarction: Looking Back to Move Forward. Focus on CMR}.
\newblock {\em Journal of Clinical Medicine}, 8(11):1805.

\bibitem[Ronneberger et~al., 2015]{ronneberger2015u}
Ronneberger, O., Fischer, P., and Brox, T. (2015).
\newblock {U-Net: Convolutional Networks for Biomedical Image Segmentation}.
\newblock In {\em Proceedings of the International Conference on Medical Image Computing and Computer-Assisted Intervention}, pages 234--241.

\bibitem[Rosenthal et~al., 1985]{rosenthal1985sudden}
Rosenthal, M.~E., Oseran, D.~S., Gang, E., and Peter, T. (1985).
\newblock {Sudden Cardiac Death following Acute Myocardial Infarction}.
\newblock {\em American Heart Journal}, 109(4):865--876.

\bibitem[Schinkel et~al., 2007]{schinkel2007assessment}
Schinkel, A.~F., Poldermans, D., Elhendy, A., and Bax, J.~J. (2007).
\newblock {Assessment of Myocardial Viability in Patients with Heart Failure}.
\newblock {\em Journal of Nuclear Medicine}, 48(7):1135--1146.

\bibitem[Selvanayagam et~al., 2004]{selvanayagam2004value}
Selvanayagam, J.~B., Kardos, A., Francis, J.~M., Wiesmann, F., Petersen, S.~E., Taggart, D.~P., and Neubauer, S. (2004).
\newblock {Value of Delayed-enhancement Cardiovascular Magnetic Resonance Imaging in Predicting Myocardial Viability after Surgical Revascularization}.
\newblock {\em Circulation}, 110(12):1535--1541.

\bibitem[Srivastava et~al., 2014]{srivastava2014dropout}
Srivastava, N., Hinton, G., Krizhevsky, A., Sutskever, I., and Salakhutdinov, R. (2014).
\newblock {Dropout: A Simple Way to Prevent Neural Networks from Overfitting}.
\newblock {\em The Journal of Machine Learning Research}, 15(1):1929--1958.

\bibitem[Van~der Wall et~al., 1996]{van1996magnetic}
Van~der Wall, E., Vliegen, H., De~Roos, A., and Bruschke, A. (1996).
\newblock {Magnetic Resonance Techniques for Assessment of Myocardial Viability}.
\newblock {\em Journal of Cardiovascular Pharmacology}, 28:37--44.

\bibitem[Walpot et~al., 2019]{walpot2019left}
Walpot, J., Juneau, D., Massalha, S., Dwivedi, G., Rybicki, F.~J., Chow, B.~J., and In{\'a}cio, J.~R. (2019).
\newblock {Left Ventricular Mid-diastolic Wall Thickness: Normal Values for Coronary CT Angiography}.
\newblock {\em Radiology: Cardiothoracic Imaging}, 1(5):e190034.

\bibitem[Wroblewski et~al., 1990]{wroblewski1990evaluation}
Wroblewski, L.~C., Aisen, A.~M., Swanson, S.~D., and Buda, A.~J. (1990).
\newblock {Evaluation of Myocardial Viability following Ischemic and Reperfusion Injury using Phosphorus 31 Nuclear Magnetic Resonance Spectroscopy in Vivo}.
\newblock {\em American Heart Journal}, 120(1):31--39.

\bibitem[Xu et~al., 2022]{xu2022bmanet}
Xu, C., Wang, Y., Zhang, D., Han, L., Zhang, Y., Chen, J., and Li, S. (2022).
\newblock {BMAnet: Boundary Mining with Adversarial Learning for Semi-supervised 2D Myocardial Infarction Segmentation}.
\newblock {\em IEEE Journal of Biomedical and Health Informatics}, 27(1):87--96.

\bibitem[Xu et~al., 2018]{xu2018direct}
Xu, C., Xu, L., Gao, Z., Zhao, S., Zhang, H., Zhang, Y., Du, X., Zhao, S., Ghista, D., Liu, H., et~al. (2018).
\newblock {Direct Delineation of Myocardial Infarction without Contrast Agents using a Joint Motion Feature Learning Architecture}.
\newblock {\em Medical Image Analysis}, 50:82--94.

\bibitem[Zabihollahy et~al., 2018]{zabihollahy2018myocardial}
Zabihollahy, F., White, J.~A., and Ukwatta, E. (2018).
\newblock {Myocardial Scar Segmentation from Magnetic Resonance Images using Convolutional Neural Network}.
\newblock In {\em Medical Imaging 2018: Computer-Aided Diagnosis}, volume 10575, pages 663--670. SPIE.

\bibitem[Zhang, 2021]{zhang2021cascaded}
Zhang, Y. (2021).
\newblock {Cascaded Convolutional Neural Network for Automatic Myocardial Infarction Segmentation from Delayed-enhancement Cardiac MRI}.
\newblock In {\em International Workshop on Statistical Atlases and Computational Models of the Heart}, pages 328--333. Springer.

\end{thebibliography}

\end{document}